\documentclass{article}

 \usepackage[dblblindworkshop, final]{neurips_2025}
\workshoptitle{UrbanAI: Harnessing Artificial Intelligence for Smart Cities}

\usepackage[utf8]{inputenc} % allow utf-8 input
\usepackage[T1]{fontenc}    % use 8-bit T1 fonts
\usepackage{hyperref}       % hyperlinks
\usepackage{url}            % simple URL typesetting
\usepackage{booktabs}       % professional-quality tables
\usepackage{amsfonts}       % blackboard math symbols
\usepackage{nicefrac}       % compact symbols for 1/2, etc.
\usepackage{microtype}      % microtypography
\usepackage{xcolor}         % colors
\usepackage{pifont}         % Added
\usepackage{multirow}       % Added
\usepackage{graphicx}
\usepackage[table]{xcolor}
\usepackage{marvosym}

\usepackage{enumitem}
\title{HeLoFusion: An Efficient and Scalable Encoder for Modeling Heterogeneous and Multi-Scale Interactions in Trajectory Prediction}

\author{
  Bingqing Wei\footnotemark[1]\quad Lianmin Chen\footnotemark[1]\hspace{0.3em}\footnotemark[2]\quad Zhongyu Xia\quad Yongtao Wang\footnotemark[3] \\
  Wangxuan Institute of Computer Technology, Peking University \\
  \texttt{bingqing.wei@stu.pku.edu.cn\quad lianminchen@hust.edu.cn}\\
  \texttt{\{xiazhongyu,wyt\}@pku.edu.cn}
}

\begin{document}
\footnotetext[1]{These authors contributed equally to this work.} % 共同一作说明
\footnotetext[2]{Lianmin Chen completed this work during his internship at the Wangxuan Institute of Computer Technology.} % 实习说明
\footnotetext[3]{Corresponding author: Yongtao Wang (\texttt{wyt@pku.edu.cn}).} % 通讯作者说明
\maketitle

\begin{abstract}
Multi-agent trajectory prediction in autonomous driving requires a comprehensive understanding of complex social dynamics. Existing methods, however, often struggle to capture the full richness of these dynamics, particularly the co-existence of multi-scale interactions and the diverse behaviors of heterogeneous agents. To address these challenges, this paper introduces HeLoFusion, an efficient and scalable encoder for modeling heterogeneous and multi-scale agent interactions. Instead of relying on global context, HeLoFusion constructs local, multi-scale graphs centered on each agent, allowing it to effectively model both direct pairwise dependencies and complex group-wise interactions (\textit{e.g.}, platooning vehicles or pedestrian crowds). Furthermore, HeLoFusion tackles the critical challenge of agent heterogeneity through an aggregation-decomposition message-passing scheme and type-specific feature networks, enabling it to learn nuanced, type-dependent interaction patterns. This locality-focused approach enables a principled representation of multi-level social context, yielding powerful and expressive agent embeddings. On the challenging Waymo Open Motion Dataset, HeLoFusion achieves state-of-the-art performance, setting new benchmarks for key metrics including Soft mAP and minADE. Our work demonstrates that a locality-grounded architecture, which explicitly models multi-scale and heterogeneous interactions, is a highly effective strategy for advancing motion forecasting.
\end{abstract}

\section{Introduction}

Trajectory prediction, forecasting the future movements of traffic participants, is crucial for autonomous driving. This task is challenging because it involves understanding complex, multi-scale social dynamics and the diverse behaviors of heterogeneous agents (\textit{e.g.}, vehicles vs. pedestrians). While many deep learning methods have been proposed, they often fall short in one of two ways. Early approaches, such as social pooling (\citet{gupta2018social, deo2018convolutional, hasan2021maneuver}), are efficient but model interactions too simplistically. Conversely, more recent methods that rely on global context, such as those using global attention (\cite{liu2021multimodal, ngiam2021scene, shi2022motion}) or dense graph neural networks (GNNs) (\citet{shi2021sgcn, xu2022groupnet}), attempt to capture all possible dependencies. However, this global view can be suboptimal, as it may dilute the influence of critical nearby agents with information from distant, irrelevant ones, while also incurring substantial computational costs. Furthermore, explicitly modeling interactions between heterogeneous agent types often leads to a combinatorial explosion in model parameters, forcing many models to treat all agents homogeneously.

To address these fundamental modeling challenges, we propose the \textbf{He}terogeneous \textbf{Lo}cal Context \textbf{Fusion} Network (\textbf{HeLoFusion}), a novel encoder designed to capture the rich structure of social interactions in a more principled manner. Our core insight is that social dynamics are inherently multi-level and type-dependent. HeLoFusion explicitly models this by constructing localized, multi-scale graphs around each agent. This allows it to capture both direct pairwise dependencies and complex group-wise interactions (\textit{e.g.}, platooning vehicles or pedestrian crowds), which are often overlooked by methods with a flat interaction structure. Furthermore, HeLoFusion tackles the critical challenge of agent heterogeneity through an aggregation-decomposition message-passing scheme and type-specific feature networks, making it learn nuanced, type-dependent interaction patterns without a prohibitive increase in parameters. In the end, by grounding these sophisticated modeling capabilities in the principle of spatial locality, our approach not only aligns with the real-world observation that an agent's behavior is primarily influenced by its immediate surroundings but also yields a highly efficient and scalable architecture.

Our key contributions are summarized as follows:
\begin{enumerate}
    \item We propose a novel graph-based module that efficiently models both \textbf{pairwise and group-wise interactions} among traffic participants by constructing multi-scale local graphs.
    \item We jointly exploit \textbf{agent heterogeneity} and \textbf{spatial locality} throughout the encoding process: category-specific networks and an aggregation–decomposition message-passing scheme operate on local neighborhoods to capture type-dependent interaction dynamics in a computationally efficient manner.
    \item Extensive experiments on the Waymo Open Motion Dataset (WOMD) demonstrate that HeLoFusion achieves \textbf{state-of-the-art performance} among comparable methods, with clear gains in Soft mAP, displacement errors, and overlap rate over strong baselines.
\end{enumerate}

\section{Related Work}
The development of large-scale datasets (\citet{ettinger2021large}) has spurred significant progress in motion prediction, with deep learning emerging as the standard approach. Early works often visualized the scene as a top-down rasterized image, leveraging convolutional networks to forecast future states (\citet{casas2020implicit, luo2021safety, zeng2021lanercnn, konev2022motioncnn}). More recently, the field has shifted towards vectorized representations, where Transformer-based architectures (\citet{jiang2023motiondiffuser, seff2023motionlm, ngiam2021scene, nayakanti2023wayformer, varadarajan2022multipath++, shi2022motion, zhou2023query}) and Graph Neural Networks (GNNs) (\citet{jia2023hdgt, mo2022multi, cui2022gorela, salzmann2020trajectron++}) are widely used to model complex social interactions. These models typically aim to capture future uncertainty by predicting either a dense occupancy grid or a sparse set of multi-modal future trajectories. State-of-the-art methods, such as MTR++ (\citet{shi2024mtr++}) and its successors (\citet{lin2024eda, sun2024controlmtr, liu2024reasoning, sun2024rmp, yan2025trajflow}), have achieved remarkable performance by capturing a global context of the scene. However, this global view can be computationally intensive and may dilute the influence of critical nearby agents. In contrast, our work is built on the principle of spatial locality, focusing on modeling structured, multi-scale interactions within local neighborhoods while explicitly addressing the heterogeneity of different agent types.

\begin{figure*}[t]
\centering
\includegraphics[width=1.0\textwidth]{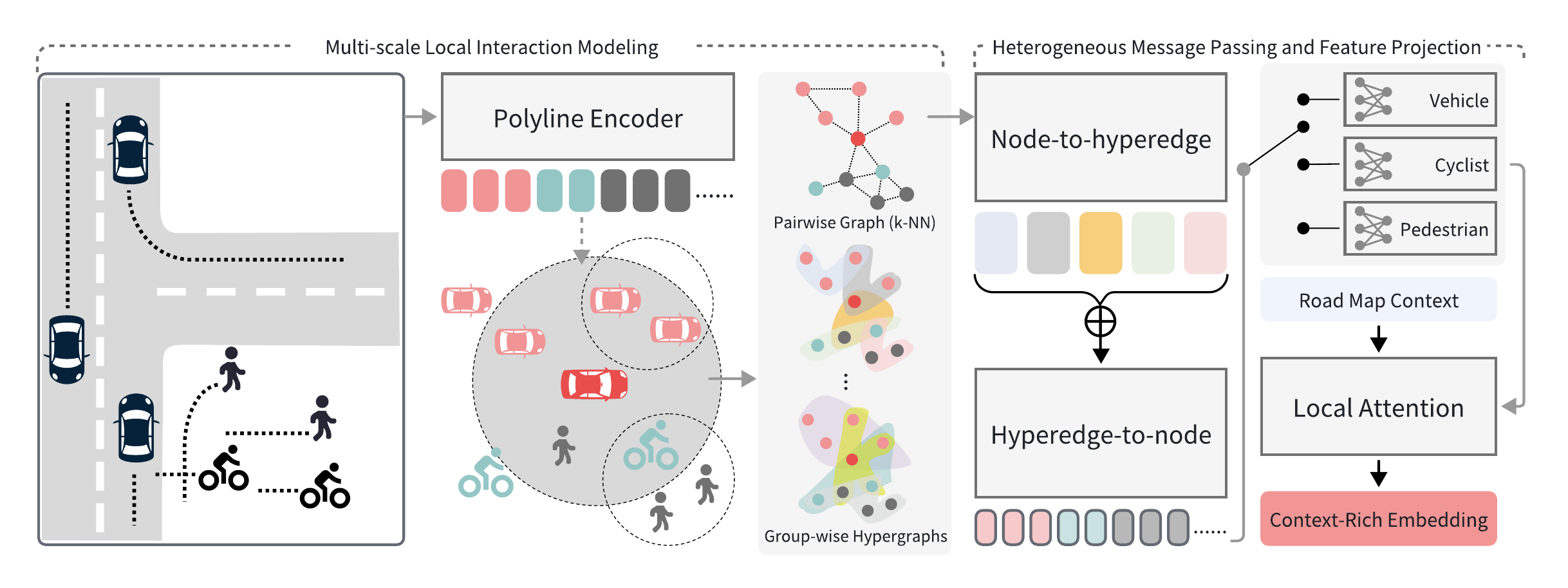}
\caption{\textbf{Overview of the HeLoFusion architecture.} The proposed encoder builds local multi-scale graphs (pairwise and group-wise) to model social interactions. It then employs a heterogeneous message-passing scheme and type-specific projection to capture the diverse dynamics of agents. Finally, a local attention module fuses the agent representations with the road map context to produce a context-rich embedding.}
\label{fig:model}
\end{figure*}

\section{Method}

Given historical trajectories $\mathcal{O} = \{ \mathbf{X}_{i}^{-} \mid i \in \mathcal{A} \}$ for all agents $\mathcal{A}$ and map information $\mathcal{M}$, our goal is to learn a rich embedding for each agent to predict its future trajectory $\hat{\mathbf{X}}_a^+$. HeLoFusion achieves this through a multi-stage encoding process that first extracts individual motion features, then models heterogeneous, multi-scale interactions, and finally fuses local scene context, as shown in Figure \ref{fig:model}.

\subsection{Capturing Pairwise and Group-wise Interactions with Local Graphs}
A core contribution of our work is the efficient modeling of complex interactions. Inspired by the spatial locality of real-world traffic scenarios, we discard expensive global graph operations and instead construct local graphs centered on each agent of interest.

\textbf{Local Graph Construction.} For each target agent, we identify its $K$ nearest neighbors. To model pairwise interactions, we construct a k-NN graph that connects these neighbors. To capture more complex group-wise interactions (\textit{e.g.}, a platoon of vehicles or a crowd of pedestrians), we construct local hypergraphs. Each hyperedge connects a neighboring agent to its own small group of nearby agents, effectively representing a local cluster. By varying the size of these groups, we create a series of graphs that capture interactions at multiple scales. This multi-scale, localized structure provides a rich yet computationally tractable foundation for modeling social dynamics.

\subsection{Modeling Agent and Interaction Heterogeneity}
Traffic participants are not homogeneous; vehicles, pedestrians, and cyclists behave and interact differently. HeLoFusion explicitly models this heterogeneity at multiple stages.

\textbf{Heterogeneous Message Passing.} To handle interactions between different agent types without a combinatorial explosion of parameters (\textit{e.g.}, separate edge types for car-pedestrian, car-cyclist, etc.), we use an aggregation-decomposition message passing scheme inspired by \citet{xu2022groupnet}. Our model first aggregates feature information from all nodes connected by an edge into a single representation. Then, a small, shared MLP dynamically decomposes this aggregated influence into category-specific messages based on the types of the participating agents. This allows interaction effects to be flexibly tailored to agent categories in a scalable manner.

\textbf{Heterogeneous Feature Projection.} To further reinforce type-specific behaviors, agent features are processed through a bank of category-specific MLPs. Each agent's embedding is routed through the MLP corresponding to its type (\textit{e.g.}, vehicle, pedestrian). This ensures that the feature representations are specialized for the distinct motion patterns and constraints of each category before being fused with scene context.

\subsection{Overall Architecture and Feature Fusion}
The principles of localized, heterogeneous interaction modeling are embedded within a three-stage encoder architecture.

\textbf{Motion Encoding.} We first process each agent's historical trajectory and all map polylines using a PointNet-style polyline encoder. This module treats a trajectory as an unordered set of points, making it robust to irregular sampling and permutation-invariant. It produces a compact feature vector for each agent and map element, with agent type embeddings included to provide initial heterogeneous signals.

\textbf{Interaction Modeling.} The motion features are then fed into the local multi-scale graph network described in Section 3.1. The heterogeneous message passing scheme (Section 3.2) is applied to produce socially-aware agent embeddings that incorporate both pairwise and group-wise dependencies.

\textbf{Context Fusion.} Finally, the interaction-aware embeddings are refined using a heterogeneous local attention module. After being projected by category-specific MLPs (Section 3.2), each agent attends to its nearby neighbors and map elements. This localized attention mechanism efficiently integrates dynamic agent information with static map constraints, producing the final context-rich embedding for the downstream prediction decoder.

\section{Experiments}

\begin{table*}[t]
    \centering
    \caption{\textbf{Performance comparison on WOMD Motion Leaderboard.} This table shows results for motion predictors without model ensemble or using extra data such as LIDAR on the {\tt{Test}} set (except for mAP-{\tt{Val}}). The best and second best results are \textbf{bolded} and \underline{underlined}, respectively.}
    \label{tab:waymo_results}
    \resizebox{\textwidth}{!}{
    \begin{tabular}{@{}l|cccc|c|cc}
        \toprule
        Method & minADE $\downarrow$ & minFDE $\downarrow$ & MR (\%) $\downarrow$ & OR (\%) $\downarrow$ & \textbf{mAP-{\tt{Val}} (\%)} $\uparrow$ & \textbf{mAP (\%)} $\uparrow$ & \textbf{Soft mAP (\%)} $\uparrow$ \\
        \midrule
        MTR++ (\citet{shi2024mtr++}) & 0.5906 & 1.1939 & 12.98 & 12.81 & 43.51 & 43.29 & 44.10 \\
        EDA (\citet{lin2024eda}) & 0.5718 & 1.1702 & 11.69 & 12.66 & 43.53 & 44.87 & 45.96 \\
        ControlMTR (\citet{sun2024controlmtr}) & 0.5897 & 1.1916 & 12.82 & \underline{12.59} & 44.13 & 44.14 & 45.72 \\
        BehaveOcc & 0.5723 & 1.1668 & 11.76 & 12.78 & - & 45.66 & 46.78 \\
        BeTopNet (\citet{liu2024reasoning}) & 0.5716 & 1.1668 & 11.83 & 12.72 & 44.16 & 45.87 & 46.98 \\
        RMP-YOLO (\citet{sun2024rmp}) & 0.5737 & 1.1697 & \textbf{11.60} & 12.66 & - & 45.23 & 46.73 \\
        TrajFlow (\citet{yan2025trajflow}) & \underline{0.5714} & \underline{1.1667} & \underline{11.62} & 12.72 & \underline{45.39} & \underline{46.04} & \underline{47.10} \\
        \textbf{HeLoFusion (Ours)} & \textbf{0.5690} & \textbf{1.1596} & 11.83 & \textbf{12.43} & \textbf{46.37} & \textbf{46.24} & \textbf{47.32} \\
        % \midrule
        % & ModeSeq & 
        \bottomrule
    \end{tabular}% 
   }
\end{table*}

\subsection{Experimental Setup}
\label{sec:exp setup}
We evaluate HeLoFusion on the large-scale Waymo Open Motion Dataset (WOMD) (\citet{ettinger2021large}), a standard benchmark for multi-agent trajectory forecasting in autonomous driving. Following the official protocol, we report key metrics including mean Average Precision (mAP), Soft mAP, minimum Average Displacement Error (minADE), minimum Final Displacement Error (minFDE), Miss Rate (MR), and Overlap Rate (OR), all evaluated with $K=6$ predictions per agent. Our model is designed as a modular encoder; for a fair and direct comparison, we replace the encoder of the BeTopNet framework (\citet{liu2024reasoning}) with HeLoFusion, while maintaining the identical decoder architecture. We compare our method against other top-performing, publicly reported methods on the WOMD leaderboard that do not use extra data such as LIDAR or ensemble techniques. More implementation details can be found in Appendix A.

\subsection{Results}
As shown in Table~\ref{tab:waymo_results}, HeLoFusion achieves state-of-the-art performance on the WOMD test set among all comparable single-model, Lidar-free methods. Our model achieves the highest Soft mAP of 47.32\% and mAP of 46.24\% on the test set; additionally, on the validation set, it also achieves the highest mAP of 46.37\%. These results demonstrate the proposed model's ability to generate accurate and well-calibrated trajectory predictions. Notably, it also achieves the lowest displacement errors (0.5690 minADE and 1.1596 minFDE) and the best Overlap Rate (12.43\%), indicating superior geometric accuracy and physical realism. Compared to its backbone, BeTopNet, HeLoFusion provides a clear improvement across most metrics, validating that our locality-focused approach effectively enhances feature representation. The strong performance confirms that modeling localized, heterogeneous interactions is a highly effective strategy for improving autonomous driving motion prediction accuracy. We also conducted a comparison of peak GPU memory usage to verify the computational scalability of our method, with more details available in Appendix \ref{sec:app:compute_resource}.

\section{Conclusion}
In this paper, we introduced HeLoFusion, a modular and efficient encoder for trajectory prediction in autonomous driving. Our approach is built on the principle of spatial locality, employing multi-scale local graphs to capture both pairwise and group-wise interactions, while explicitly handling agent heterogeneity, thereby producing powerful, socially aware agent representations. Our state-of-the-art results on the WOMD validate that the proposed design is a highly effective and practical strategy, offering an efficient module suitable for real-world autonomous driving systems.

\begin{ack}
This work was supported by National Key R\&D Program of China (Grant No.2022ZD0160305).
\end{ack}

{
\small
\bibliographystyle{plainnat}
\bibliography{bib/final}

\begin{thebibliography}{28}
\providecommand{\natexlab}[1]{#1}
\providecommand{\url}[1]{\texttt{#1}}
\expandafter\ifx\csname urlstyle\endcsname\relax
  \providecommand{\doi}[1]{doi: #1}\else
  \providecommand{\doi}{doi: \begingroup \urlstyle{rm}\Url}\fi

\bibitem[Casas et~al.(2020)Casas, Gulino, Suo, Luo, Liao, and
  Urtasun]{casas2020implicit}
Sergio Casas, Cole Gulino, Simon Suo, Katie Luo, Renjie Liao, and Raquel
  Urtasun.
\newblock Implicit latent variable model for scene-consistent motion
  forecasting.
\newblock In \emph{European Conference on Computer Vision}, pages 624--641.
  Springer, 2020.

\bibitem[Cui et~al.(2022)Cui, Casas, Wong, Suo, and Urtasun]{cui2022gorela}
Alexander Cui, Sergio Casas, Kelvin Wong, Simon Suo, and Raquel Urtasun.
\newblock Gorela: Go relative for viewpoint-invariant motion forecasting.
\newblock \emph{arXiv preprint arXiv:2211.02545}, 2022.

\bibitem[Deo and Trivedi(2018)]{deo2018convolutional}
Nachiket Deo and Mohan~M Trivedi.
\newblock Convolutional social pooling for vehicle trajectory prediction.
\newblock In \emph{Proceedings of the IEEE conference on computer vision and
  pattern recognition workshops}, pages 1468--1476, 2018.

\bibitem[Ettinger et~al.(2021)Ettinger, Cheng, Caine, Liu, Zhao, Pradhan, Chai,
  Sapp, Qi, Zhou, et~al.]{ettinger2021large}
Scott Ettinger, Shuyang Cheng, Benjamin Caine, Chenxi Liu, Hang Zhao, Sabeek
  Pradhan, Yuning Chai, Ben Sapp, Charles~R Qi, Yin Zhou, et~al.
\newblock Large scale interactive motion forecasting for autonomous driving:
  The waymo open motion dataset.
\newblock In \emph{Proceedings of the IEEE/CVF international conference on
  computer vision}, pages 9710--9719, 2021.

\bibitem[Gupta et~al.(2018)Gupta, Johnson, Fei-Fei, Savarese, and
  Alahi]{gupta2018social}
Agrim Gupta, Justin Johnson, Li~Fei-Fei, Silvio Savarese, and Alexandre Alahi.
\newblock Social gan: Socially acceptable trajectories with generative
  adversarial networks.
\newblock In \emph{Proceedings of the IEEE conference on computer vision and
  pattern recognition}, pages 2255--2264, 2018.

\bibitem[Hasan et~al.(2021)Hasan, Solernou, Paschalidis, Wang, Markkula, and
  Romano]{hasan2021maneuver}
Mohamed Hasan, Albert Solernou, Evangelos Paschalidis, He~Wang, Gustav
  Markkula, and Richard Romano.
\newblock Maneuver-aware pooling for vehicle trajectory prediction.
\newblock \emph{arXiv preprint arXiv:2104.14079}, 2021.

\bibitem[Jia et~al.(2023)Jia, Wu, Chen, Liu, Li, and Yan]{jia2023hdgt}
Xiaosong Jia, Penghao Wu, Li~Chen, Yu~Liu, Hongyang Li, and Junchi Yan.
\newblock Hdgt: Heterogeneous driving graph transformer for multi-agent
  trajectory prediction via scene encoding.
\newblock \emph{IEEE transactions on pattern analysis and machine
  intelligence}, 45\penalty0 (11):\penalty0 13860--13875, 2023.

\bibitem[Jiang et~al.(2023)Jiang, Cornman, Park, Sapp, Zhou, Anguelov,
  et~al.]{jiang2023motiondiffuser}
Chiyu Jiang, Andre Cornman, Cheolho Park, Benjamin Sapp, Yin Zhou, Dragomir
  Anguelov, et~al.
\newblock Motiondiffuser: Controllable multi-agent motion prediction using
  diffusion.
\newblock In \emph{Proceedings of the IEEE/CVF conference on computer vision
  and pattern recognition}, pages 9644--9653, 2023.

\bibitem[Konev et~al.(2022)Konev, Brodt, and Sanakoyeu]{konev2022motioncnn}
Stepan Konev, Kirill Brodt, and Artsiom Sanakoyeu.
\newblock Motioncnn: A strong baseline for motion prediction in autonomous
  driving.
\newblock \emph{arXiv preprint arXiv:2206.02163}, 2022.

\bibitem[Lin et~al.(2024)Lin, Lin, Lin, Huang, Xiong, and Wang]{lin2024eda}
Longzhong Lin, Xuewu Lin, Tianwei Lin, Lichao Huang, Rong Xiong, and Yue Wang.
\newblock Eda: Evolving and distinct anchors for multimodal motion prediction.
\newblock In \emph{Proceedings of the AAAI Conference on Artificial
  Intelligence}, volume~38, pages 3432--3440, 2024.

\bibitem[Liu et~al.(2024)Liu, Chen, Qiao, Lv, and Li]{liu2024reasoning}
Haochen Liu, Li~Chen, Yu~Qiao, Chen Lv, and Hongyang Li.
\newblock Reasoning multi-agent behavioral topology for interactive autonomous
  driving.
\newblock \emph{Advances in Neural Information Processing Systems},
  37:\penalty0 92605--92637, 2024.

\bibitem[Liu et~al.(2021)Liu, Zhang, Fang, Jiang, and Zhou]{liu2021multimodal}
Yicheng Liu, Jinghuai Zhang, Liangji Fang, Qinhong Jiang, and Bolei Zhou.
\newblock Multimodal motion prediction with stacked transformers.
\newblock In \emph{Proceedings of the IEEE/CVF conference on computer vision
  and pattern recognition}, pages 7577--7586, 2021.

\bibitem[Luo et~al.(2021)Luo, Casas, Liao, Yan, Xiong, Zeng, and
  Urtasun]{luo2021safety}
Katie Luo, Sergio Casas, Renjie Liao, Xinchen Yan, Yuwen Xiong, Wenyuan Zeng,
  and Raquel Urtasun.
\newblock Safety-oriented pedestrian occupancy forecasting.
\newblock In \emph{2021 IEEE/RSJ International Conference on Intelligent Robots
  and Systems (IROS)}, pages 1015--1022. IEEE, 2021.

\bibitem[Mo et~al.(2022)Mo, Huang, Xing, and Lv]{mo2022multi}
Xiaoyu Mo, Zhiyu Huang, Yang Xing, and Chen Lv.
\newblock Multi-agent trajectory prediction with heterogeneous edge-enhanced
  graph attention network.
\newblock \emph{IEEE Transactions on Intelligent Transportation Systems},
  23\penalty0 (7):\penalty0 9554--9567, 2022.

\bibitem[Nayakanti et~al.(2023)Nayakanti, Al-Rfou, Zhou, Goel, Refaat, and
  Sapp]{nayakanti2023wayformer}
Nigamaa Nayakanti, Rami Al-Rfou, Aurick Zhou, Kratarth Goel, Khaled~S Refaat,
  and Benjamin Sapp.
\newblock Wayformer: Motion forecasting via simple \& efficient attention
  networks.
\newblock In \emph{2023 IEEE International Conference on Robotics and
  Automation (ICRA)}, pages 2980--2987. IEEE, 2023.

\bibitem[Ngiam et~al.(2021)Ngiam, Caine, Vasudevan, Zhang, Chiang, Ling,
  Roelofs, Bewley, Liu, Venugopal, et~al.]{ngiam2021scene}
Jiquan Ngiam, Benjamin Caine, Vijay Vasudevan, Zhengdong Zhang, Hao-Tien~Lewis
  Chiang, Jeffrey Ling, Rebecca Roelofs, Alex Bewley, Chenxi Liu, Ashish
  Venugopal, et~al.
\newblock Scene transformer: A unified architecture for predicting multiple
  agent trajectories.
\newblock \emph{arXiv preprint arXiv:2106.08417}, 2021.

\bibitem[Salzmann et~al.(2020)Salzmann, Ivanovic, Chakravarty, and
  Pavone]{salzmann2020trajectron++}
Tim Salzmann, Boris Ivanovic, Punarjay Chakravarty, and Marco Pavone.
\newblock Trajectron++: Dynamically-feasible trajectory forecasting with
  heterogeneous data.
\newblock In \emph{European Conference on Computer Vision}, pages 683--700.
  Springer, 2020.

\bibitem[Seff et~al.(2023)Seff, Cera, Chen, Ng, Zhou, Nayakanti, Refaat,
  Al-Rfou, and Sapp]{seff2023motionlm}
Ari Seff, Brian Cera, Dian Chen, Mason Ng, Aurick Zhou, Nigamaa Nayakanti,
  Khaled~S Refaat, Rami Al-Rfou, and Benjamin Sapp.
\newblock Motionlm: Multi-agent motion forecasting as language modeling.
\newblock In \emph{Proceedings of the IEEE/CVF International Conference on
  Computer Vision}, pages 8579--8590, 2023.

\bibitem[Shi et~al.(2021)Shi, Wang, Long, Zhou, Zhou, Niu, and
  Hua]{shi2021sgcn}
Liushuai Shi, Le~Wang, Chengjiang Long, Sanping Zhou, Mo~Zhou, Zhenxing Niu,
  and Gang Hua.
\newblock Sgcn: Sparse graph convolution network for pedestrian trajectory
  prediction.
\newblock In \emph{Proceedings of the IEEE/CVF conference on computer vision
  and pattern recognition}, pages 8994--9003, 2021.

\bibitem[Shi et~al.(2022)Shi, Jiang, Dai, and Schiele]{shi2022motion}
Shaoshuai Shi, Li~Jiang, Dengxin Dai, and Bernt Schiele.
\newblock Motion transformer with global intention localization and local
  movement refinement.
\newblock \emph{Advances in Neural Information Processing Systems},
  35:\penalty0 6531--6543, 2022.

\bibitem[Shi et~al.(2024)Shi, Jiang, Dai, and Schiele]{shi2024mtr++}
Shaoshuai Shi, Li~Jiang, Dengxin Dai, and Bernt Schiele.
\newblock Mtr++: Multi-agent motion prediction with symmetric scene modeling
  and guided intention querying.
\newblock \emph{IEEE Transactions on Pattern Analysis and Machine
  Intelligence}, 46\penalty0 (5):\penalty0 3955--3971, 2024.

\bibitem[Sun et~al.(2024{\natexlab{a}})Sun, Li, Liu, Yuan, Sun, Huang, Wong,
  Tee, and Ang~Jr]{sun2024rmp}
Jiawei Sun, Jiahui Li, Tingchen Liu, Chengran Yuan, Shuo Sun, Zefan Huang,
  Anthony Wong, Keng~Peng Tee, and Marcelo~H Ang~Jr.
\newblock Rmp-yolo: A robust motion predictor for partially observable
  scenarios even if you only look once.
\newblock \emph{arXiv preprint arXiv:2409.11696}, 2024{\natexlab{a}}.

\bibitem[Sun et~al.(2024{\natexlab{b}})Sun, Yuan, Sun, Wang, Han, Ma, Huang,
  Wong, Tee, and Ang]{sun2024controlmtr}
Jiawei Sun, Chengran Yuan, Shuo Sun, Shanze Wang, Yuhang Han, Shuailei Ma,
  Zefan Huang, Anthony Wong, Keng~Peng Tee, and Marcelo~H Ang.
\newblock Controlmtr: Control-guided motion transformer with scene-compliant
  intention points for feasible motion prediction.
\newblock In \emph{2024 IEEE 27th International Conference on Intelligent
  Transportation Systems (ITSC)}, pages 1507--1514. IEEE, 2024{\natexlab{b}}.

\bibitem[Varadarajan et~al.(2022)Varadarajan, Hefny, Srivastava, Refaat,
  Nayakanti, Cornman, Chen, Douillard, Lam, Anguelov,
  et~al.]{varadarajan2022multipath++}
Balakrishnan Varadarajan, Ahmed Hefny, Avikalp Srivastava, Khaled~S Refaat,
  Nigamaa Nayakanti, Andre Cornman, Kan Chen, Bertrand Douillard, Chi~Pang Lam,
  Dragomir Anguelov, et~al.
\newblock Multipath++: Efficient information fusion and trajectory aggregation
  for behavior prediction.
\newblock In \emph{2022 International Conference on Robotics and Automation
  (ICRA)}, pages 7814--7821. IEEE, 2022.

\bibitem[Xu et~al.(2022)Xu, Li, Ni, Zhang, and Chen]{xu2022groupnet}
Chenxin Xu, Maosen Li, Zhenyang Ni, Ya~Zhang, and Siheng Chen.
\newblock Groupnet: Multiscale hypergraph neural networks for trajectory
  prediction with relational reasoning.
\newblock In \emph{Proceedings of the IEEE/CVF Conference on Computer Vision
  and Pattern Recognition}, pages 6498--6507, 2022.

\bibitem[Yan et~al.(2025)Yan, Zhang, Zhang, Yang, White, Chen, Liu, Liu,
  Zhuang, Shi, et~al.]{yan2025trajflow}
Qi~Yan, Brian Zhang, Yutong Zhang, Daniel Yang, Joshua White, Di~Chen, Jiachao
  Liu, Langechuan Liu, Binnan Zhuang, Shaoshuai Shi, et~al.
\newblock Trajflow: Multi-modal motion prediction via flow matching.
\newblock \emph{arXiv preprint arXiv:2506.08541}, 2025.

\bibitem[Zeng et~al.(2021)Zeng, Liang, Liao, and Urtasun]{zeng2021lanercnn}
Wenyuan Zeng, Ming Liang, Renjie Liao, and Raquel Urtasun.
\newblock Lanercnn: Distributed representations for graph-centric motion
  forecasting.
\newblock In \emph{2021 IEEE/RSJ International Conference on Intelligent Robots
  and Systems (IROS)}, pages 532--539. IEEE, 2021.

\bibitem[Zhou et~al.(2023)Zhou, Wang, Li, and Huang]{zhou2023query}
Zikang Zhou, Jianping Wang, Yung-Hui Li, and Yu-Kai Huang.
\newblock Query-centric trajectory prediction.
\newblock In \emph{Proceedings of the IEEE/CVF conference on computer vision
  and pattern recognition}, pages 17863--17873, 2023.

\end{thebibliography}
}

%%%%%%%%%%%%%%%%%%%%%%%%%%%%%%%%%%%%%%%%%%%%%%%%%%%%%%%%%%%%

\appendix

\section{Implementation Details}
\label{sec:app:inp details}

\textbf{Framework and Architecture.} Our model is built upon the BeTopNet framework (\citet{liu2024reasoning}), replacing its original encoder with our proposed HeLoFusion module while keeping the decoder architecture unchanged for a fair comparison. Following the protocol of MTR (\cite{shi2022motion}), we use 64 predefined intention points as prediction anchors, which are generated via k-means clustering on the training set. Within the HeLoFusion encoder, the multi-scale interaction module constructs pairwise graphs from the $K=10$ nearest neighbors and group-wise interactions using two hypergraphs with hyperedge sizes $S^{(1)}$ of $5$ and $7$. The subsequent context fusion module uses a local attention mechanism with a neighborhood size of 16. To account for agent heterogeneity, we employ three category-specific MLPs for the vehicle, pedestrian, and cyclist classes in the WOMD dataset.

\textbf{Training.} The model is trained for 30 epochs on 8 NVIDIA GeForce RTX 3090 GPUs. We use the AdamW optimizer with an initial learning rate of $1 \times 10^{-4}$ and a weight decay of $0.01$. The learning rate is managed by a step scheduler, decaying by a factor of $0.5$ at epochs 22, 24, 26, and 28. We use a batch size of 80 and apply gradient accumulation to accommodate memory constraints. For training stability, gradient clipping is applied with a maximum norm of $1000.0$. The network is initialized with weights from a BeTopNet model pretrained for 30 epochs.

\section{Computational Resource Requirement Analysis}
\label{sec:app:compute_resource}

To verify the computational scalability of HeLoFusion enabled by its spatial locality design, we compare the peak GPU memory usage during training between our method that leverages spatial locality and the global method that does not exploit spatial locality. All other experimental settings are kept identical between the two methods. In the experiment, we fix the number of target agents to predict at 4, systematically vary the total number of traffic participants in the scene, and measure the peak GPU memory usage of both methods during training.

\begin{table*}[t]
    \centering
    \caption{\textbf{Comparison of Peak GPU Memory Usage During Training (Unit: MB)}. The ``w/ Locality'' column corresponds to the memory consumption of HeLoFusion (with spatial locality), and the ``w/o Locality'' column corresponds to that of the global method without spatial locality.}
    \label{tab:memory_usage}
    \begin{tabular}{ccc@{}}
        \toprule
        \#Traffic Participants & w/ Locality & w/o Locality \\
        \midrule
        29 & 2079.67 & 2166.51 \\
        92 & 2170.80 & 4918.78 \\
        122 & 2109.16 & 9984.98 \\
        165 & 2600.44 & 23223.32 \\
        235 & 2875.38 & 64858.23 \\
        \bottomrule
    \end{tabular}
\end{table*}

As shown in Table~\ref{tab:memory_usage}, as the number of traffic participants in the scene increases, the peak memory of the global method without spatial locality grows exponentially, while the memory consumption of HeLoFusion exhibits a significantly more moderate growth trend. This result verifies that our spatial locality modeling achieves scalability in memory usage. Even in complex traffic scenarios with massive participants, HeLoFusion maintains low computational resource demands—a critical advantage for the practical deployment of autonomous driving systems. 

\section{Limitations}
\label{sec:app:limit}
Our work, while promising, has certain limitations. The proposed model has been validated on the Waymo Open Motion Dataset; its generalization to different datasets with unique traffic patterns and environmental conditions remains to be explored. And like other data-driven methods, its performance on rare, out-of-distribution events is not guaranteed.

\section{Broader Impacts Statement}
\label{sec:app:impacts}
The primary positive societal impact of our research is its potential to enhance autonomous driving safety, leading to fewer traffic accidents and improved transportation efficiency. However, we acknowledge potential negative consequences. The advancement of autonomous systems raises ethical questions regarding decision-making in unavoidable collisions. It is crucial that the development of such technology is paired with rigorous safety validation, research into algorithmic fairness to prevent biases, and thoughtful public policy to manage its societal transition.

%%%%%%%%%%%%%%%%%%%%%%%%%%%%%%%%%%%%%%%%%%%%%%%%%%%%%%%%%%%%

% \input{checklist}

\end{document}